\documentclass[11pt,a4paper]{article}
\usepackage[hyperref]{acl2021}
\usepackage{acl2021}
\usepackage{times}
\usepackage{latexsym}

\usepackage{microtype}
\usepackage{amsfonts}
\usepackage{amsmath}
\usepackage{amssymb}
\usepackage{breqn}
\usepackage{graphicx}
\usepackage{tikz}
\usepackage{xspace}
\usepackage{xcolor}
\usepackage{lipsum}
\usepackage{makecell}
\usepackage{tabularx}
\usepackage{listings}
\usepackage{textcomp}
\usepackage{color, colortbl}
\usepackage{fancyvrb}
\usepackage[symbol]{footmisc}

\definecolor{light-gray}{gray}{0.95}
\usepackage[shortlabels]{enumitem}

\newcommand{\x}{\pmb{x}}
\newcommand{\y}{\pmb{y}}
\newcommand{\vtheta}{\pmb{\theta}}

\usepackage{microtype}

\aclfinalcopy % Uncomment this line for the final submission
\title{Code Generation from Natural Language with \\
 Less Prior and More Monolingual Data}

\author{Sajad Norouzi\footnotemark ~~~~   Keyi Tang~~~~  Yanshuai Cao \\
  Borealis AI\\
  \small \texttt{sajadn@cs.toronto.edu,\{keyi.tang,yanshuai.cao\}@borealisai.com} \\
}

\date{}

\begin{document}
\maketitle
\footnotetext[1]{Work done during internship at BorealisAI}
\begin{abstract}
  %Semantic parsing is the task of converting natural language utterances to machine-understandable meaning representations, such as logic forms or programming languages.

Training datasets for semantic parsing are typically small due to the higher expertise required for annotation than most other NLP tasks. As a result, models for this application usually need additional prior knowledge to be built into the architecture or algorithm. The increased dependency on human experts hinders automation and raises the development and maintenance costs in practice. This work investigates whether a generic transformer-based seq2seq model can achieve competitive performance with minimal code-generation-specific inductive bias design. By exploiting a relatively sizeable monolingual corpus of the target programming language, which is cheap to mine from the web, we achieved $81.03\%$ exact match accuracy on Django and $32.57$ BLEU score on CoNaLa. Both are SOTA to the best of our knowledge. This positive evidence highlights a potentially easier path toward building accurate semantic parsers in practice. \footnote[2]{Code at \href{https://github.com/BorealisAI/code-gen-TAE}{https://github.com/BorealisAI/code-gen-TAE}}

%% \sajad{to verify SOTA}

%% In this work, we present a simple sequential natural language to programming language translator without any prior knowledge about the target programming language. In contrast to recent approaches, we don't utilize any pre-specified grammar for the code generation simplifying the architecture significantly. We also demonstrate how to incorporate monolingual data from programming languages to improve the model. Experiments indicates the effectiveness of our approach reaching state of the art results on python datasets. We achieved 80.78\% exact match accuracy on Django and 32.29 BELU score on CoNaLa.

%% In this work, we present a sequential natural language to programming language translator. In contrast to popular approaches, we don't utilize a pre-specified grammar for the code generation simplifying the architecture. We also prepose incorporating monolingual data from programming languages to improve the model. Experiments demonstrate the effectiveness of our approach reaching state of the art numbers on python datasets. We achieved 80.78\% exact match accuracy on Django and 32.29 BELU score on CoNaLa.

%% In this work, we present a sequential natural language to programming language translator. In contrast to popular approaches, we don't utilize a pre-specified grammar for the code generation simplifying the architecture. We also prepose incorporating monolingual data from programming languages to improve the model. 

\end{abstract}

\vspace{-0.15cm}
\section{Introduction}
\vspace{-0.15cm}

%\blfootnote{Work in progress}
For a machine to act upon users' natural language inputs, a model needs to convert the natural language utterances to machine-understandable meaning representation, i.e. semantic parsing (SP). The output meaning representation is beyond shallow identification of topic, intention, entity or relation, but complex structured objects expressed as logical forms, query language or general-purpose programs. Therefore, annotating parallel corpus for semantic parsing requires more costly expertise. 

\begin{figure}[h]
	\begin{center}
	\includegraphics[width=.85\linewidth]{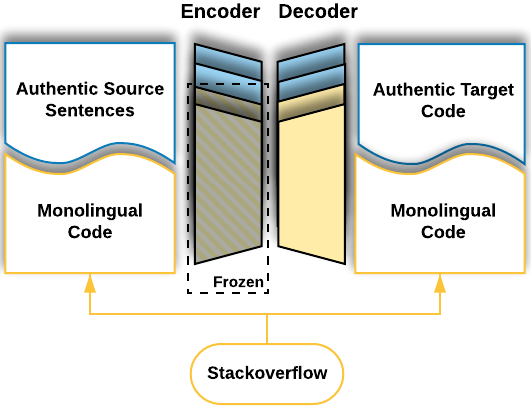}
	\caption{TAE: the monolingual corpus is used both as source and target. The encoder is frozen in the computation branch on the monolingual data.}
	\label{fig:system}
	\end{center}
\vspace{-0.6cm}
\end{figure}

SP shares some resemblance with machine translation (MT). However, SP datasets are typically smaller, with only a few thousand to at most tens of thousands of examples, even smaller than most low resource MT problems. Simultaneously, because the predicted outputs generally need to be exactly correct to execute and produce the right answer, the accuracy requirement is generally higher than MT. As a result, inductive bias design in architecture and algorithm has been prevalent in the SP literature \cite{dong2016language,yin2017syntactic,yin2018tranx,dong2018coarse,guo2019towards,wang2019rat,yin2019reranking}.

%% and has produced significant improvements over vanilla sequential models typically used in other sequence prediction NLP tasks. 

While their progress is remarkable, excessive task-specific expert design makes the models complicated, hard to transfer to new domains, and challenging to deploy in real-world applications. In this work, we look at the opposite end of the spectrum and try to answer the following question: {\em with little inductive bias in the model, and no additional labelled data, is it still possible to achieve competitive performance?} This is an important question, as the answer could point to a much shorter road to practical SP without breaking the bank. 

This paper shows that the answer is encouragingly affirmative.  By exploiting a relatively large monolingual corpus of the programming language, a transformer-based Seq2Seq model \cite{vaswani2017attention} with little SP specific prior could potentially attain results superior to or competitive with the state-of-the-art models specially designed for semantic parsing. %% Furthermore, the monolingual corpus only needs to be in the same meaning representation language and does not have to come from the same distribution, hence in practice, it is possible to mine even larger quantities of such examples from the web.
Our contributions are three-fold:
\begin{itemize}
\item We provide evidence that transformer-based seq2seq models can reach a competitive or superior performance with models specifically designed for semantic parsing. This suggests an alternative route for future progress other than inductive bias design;
\item We do empirical analysis over previously proposed approaches for incorporating monolingual data and show the effectiveness of our modified technique on a range of datasets;
% We show how to use monolingual corpora effectively for SP;
% , as well as a dataset metric to judge when the additional monolingual might improve the performance and when it might not; %% and that not all techniques for using monolingual corpra from machine translation carry over to this task.
%\item We show when the additional monolingual dataset might improve the performance and when it might not;
\item We set the new state-of-the-art on Django \cite{oda2015ase:pseudogen1} reaching 81.03\% exact match accuracy and on CoNaLa \cite{yin2018mining} with a BLEU score of 32.57.

\end{itemize}

\section{Previous Work on Semantic Parsing}
\label{sec:related}
Different sources of prior knowledge about the SP problem structure could be exploited.

%% On the high-level, works could design model inductive biases on the input encoding \cite{wang2019rat,herzig2020span}, on the output decoding \cite{dong2016language,yin2017syntactic,yin2018tranx,dong2018coarse,guo2019towards}, on the reasoning steps that connect inputs to outputs \cite{zelle1996learning,zettlemoyer2007online,clarke-etal-2010-driving,liang2013learning,herzig2020span}, or by designing task-specific pre-training strategies that leverage knowledge about the data generation process \cite{herzig2020tapas,yu2020grappa}.

\noindent \textbf{Input structure:} \citet{wang2019rat} adapts the transformer relative position encoding \cite{shaw2018self} to express relations among the database schema elements as well as with the input text spans. \citet{herzig2020span} proposed a span-based neural parser with compositional inductive bias built-in. \citet{herzig2020span} also leverages a CKY-style \cite{cocke1969programming,kasami1966efficient,younger1967recognition} inference to link input features to output codes.
% which have been used in SP long before the modern neural approaches \cite{zelle1996learning,zettlemoyer2007online,clarke-etal-2010-driving,liang2013learning}.

\noindent \textbf{Output structure:} The implicit tree or graph-like structures in the programs can also be exploited. \citet{dong2016language} proposed parent-feeding LSTM following the tree structure. \citet{dong2018coarse} proposed a coarse-to-fine decoding approach. \citet{guo2019towards} crafted an intermediate meaning representation to bridge the large gap between input utterance and the output SQL queries. \citet{yin2017syntactic,yin2018tranx} proposed TranX, a more general-purpose transition-based system, to ensure grammaticality of predictions. Using TranX, the neural model predicts the linear sequence of AST-tree constructing actions instead of the program tokens. However, a human expert needs to craft the grammar, and the design quality impacts the learning and generalization for the neural nets.

% \noindent \textbf{Data:} \citet{yu2020grappa} leverages strong human expertise by constructing a synchronous context-free grammar that can generate synthetic question-SQL pairs for task-specific supervised pre-training.

Sequential models with less SP specific priors have been investigated \cite{dong2016language, ling2016latent, Zeng2020PhotonAR}, However, they generally fell short in accuracy comparing to the best of structure-exploiting ones listed above. %% Nonetheless, a more generic architecture remains appealing: the ease of deployment and maintenance in practice thanks to less reliance on human expertise; and the ease of transfer to other applications thanks to improved modularity. Therefore, we would like to explore whether the current best transformer-based seq2seq architecture can achieve competitive results.

%% Grammar based approaches convert the code into an abstract syntax tree (AST) and define a set of grammar rules to generate ASTs \cite{yin2018tranx, guo2019towards}. This approach guarantees that the generated sequences are grammatically correct and embedding this prior knowledge probably help the model to hunger for less data. In contrast, this approach makes the architecture complicated. More importantly, the design of a good grammar requires human effort and we cannot make sure that the designed grammar is optimal. This process should be repeated for each target language from scratch which makes it a time consuming procedure. 
%% \par
%% We believe one reason behind the success of grammar based approaches is the small size of semantic parsing datasets which makes it a proper area for tailoring different type of inductive biases. At the same time, annotation of programming languages can be expensive, that's why we try to leverage unlabeled data. Note that on some domains like SQL gathering unlabeled datasets is not straight forward either. 
%% \par
The most closely related to ours is the work by \citet{xu2020incorporating} for incorporating external knowledge from extra datasets, which used a noisy parallel dataset from Stackoverflow to pre-train the SP and fine-tuned it on the primary dataset. Their approach's main limitation is still the need for (noisy) parallel data, albeit cheaper than the primary labelled set. Nonetheless, as we shall see in the experiment section later, our approach achieves better results when using the same amount of data mined from the same source despite ignoring the source sentence.

\begin{figure*}[!h]
    \centering
    \includegraphics[width=\linewidth]{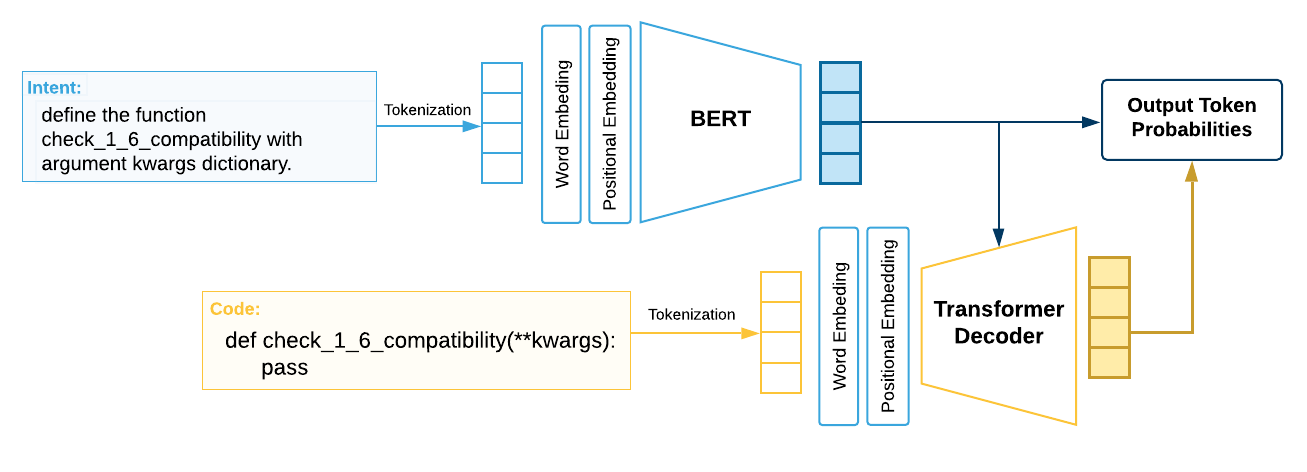}
    \caption{Model overview during training: we use a standard transformer-based encoder-decoder model where the positional and word embeddings are shared between encoder and decoder. The modules related to the encoder are represented in blue and the decoder ones are in yellow. Standard teacher forcing and transformer masking is applied during training.}
    %% \caption{\label{fig:architecture}The overview of the architecture. We used a transformer based encoder-decoder model where the positional and word embeddings are share between encoder and decoder. The logits are computed based on equation \ref{eq:logits}.}
    \label{fig:architecture}
\end{figure*}
%% =======
%% \sajad{\citet{xu2020incorporating} used noisy parallel dataset from Stackoverflow to pre-train the SP and fine-tune it on the primary dataset. They also proposed a method to sample noisy pairs similar to the primary bitext more. The main limitation of this approach is that we need to have access to parallel data which limits the amount of data we can use. Here we focus on monolingual data instead.}   
%% >>>>>>> db9a10ae3f8b9ccd5c26798661ea8353d498c0da

\section{Background and Methodology}

BERT \cite{devlin2018bert} class of pre-trained models can make up for the lack of inductive bias on the input side to some degree. On the output side, we hope to learn the necessary prior knowledge about the target meaning representation from unlabelled monolingual data.

%\nointent \textbf{Background:}
Using monolingual data to improve seq2seq models is not new and has been extensively studied in MT before. Notable methods include {\em fusion} \cite{gulcehre2015using,ramachandran2016unsupervised,Sriram2018ColdFT,Stahlberg2018SimpleFR}, {\em back-translation (BT)} \cite{sennrich2015improving,edunov2018understanding,hoang2018iterative}, \cite{currey2017copied, Burlot2018UsingMD,burlot2019using}, and {\em BT with copied monolingual data} \cite{currey2017copied, burlot2019using}. However, due to more structured outputs, less training data, and different evaluation metrics of exact match correctness instead of BLEU, it is unclear if these lessons transfer from MT to SP. So SP-specific investigation is needed.

%% \subsection{Approaches to Use Monolingual Data}
%% \label{sec:background}

%% \sajad {The motivation behind this idea is that several source tokens need to be kept in the target sentence, so they added an auxiliary term to make sure target sentences can be reconstructed correctly. This is the case for semantic parsing where function or attribute names need to be preserved in the target sentence.}

%% \simon{I think you could improve this paragraph.  It's important to say that they added an extra term to the objective function that makes sure that target sentences are copied across correctly.   It would be good to give this a name throughout the paper (target autoencoding objective?) as it is easily confused with the copy-attention mechanism if you don't know much about this.  Also, what is the intuition behind this method?  At the moment you just motivate it based on the failures of back-translation.} 

%% \begin{equation}
%%     \mathcal{L} = E_{p(\x_i, \y_i)} ~ T_{\vtheta}(\y_i \mid \x_i) ~ + ~ \lambda E_{p(\y')} ~ L_{\vtheta}(\y')
%%     \label{eq:objective}
%% \end{equation}
%% Here we augmented the conventional translation objective with a language model (LM) on a monolingual dataset. $\lambda$ controls the balance between the supervised and monolingual sets. 

%% \input{background}
%% \subsection{Architecture}

\subsection{Target Autoencoding \small{with Frozen Encoder}}
\label{sec:main_method}

% \sajad{this paragraph doesn't fit into the story. In the previous paragraph we said use of monolingual data in SP requires epxeriments and here we say we follow others on copy target}
% \citet{currey2017copied,burlot2019using} observed that in low resource MT, learning by auto-encoding the monolingual data beside the main supervised training is more effective than all the variants of back-translation and fusion. Hence our method is based on this approach. 

% Due to the gap between natural language and programming language, if there are similar improvement in SP, the benefits would likely be in helping the decoder training. Hence, we conjecture that one needs to freeze the encoder on the auto-encoding part of the computational graph, as illustrated in Fig.\ \ref{fig:system}. The ablation study in Sec.\ \ref{sec:ablation} confirms this hypothesis later.

% Here we propose Target Auto-encoding, a simple variant of coppied monolingual data and in section 4.1 we do ablation study over different approaches. 

We assume a parallel corpus of natural language utterances and their corresponding programs, $\mathcal{B}=\{\x_i, \y_i\}$.
The goal is to train a translator model (TM) to maximize the conditional log probability of $\y_i$ given $\x_i$, $T_{\vtheta}(\y_i|\x_i)$, over the training set: $\mathcal{L}_{\text{sup}} = {\textstyle \sum}_{\mathcal{B}} T_{\vtheta}(\y_i \mid \x_i)$
%% \begin{equation}    
%%     \label{eq:objective_conv}
%% \end{equation}
\noindent where $\vtheta$ is the vector of TM model parameters. Let $\mathcal{M}=\{\y'_i\}$ denote the monolingual dataset in the target language.
\par
\citet{currey2017copied,burlot2019using} demonstrated that in low resource MT, auto-encoding the monolingual data besides the main supervised training is helpful. Following the same path, we add an auto-encoding objective term on monolingual data:
$\mathcal{L_{\text{full}}} = \mathcal{L}_{\text{sup}} + {\textstyle\sum}_{\mathcal{M}} T_{\vtheta}(\y'_i \mid \y'_i)$.
%% \begin{equation}
%%     \mathcal{L_{\text{full}}} = \mathcal{L}_{\text{sup}} + {\textstyle\sum}_{\mathcal{M}} T_{\vtheta}(\y'_i \mid \y'_i) \label{eq:copy_obj}
%% \end{equation}
The target $\y'_i$'s are reconstructed using the shared encoder-decoder model.
\par
We conjecture that monolingual data auto-encoding mainly helps the decoder, so we propose to freeze the encoder parameters for monolingual data.
Writing the encoder and decoder parameters separately with $\vtheta = [\vtheta_e, \vtheta_d]$, then $\vtheta_e$ is updated using the gradient of the supervised objective $\mathcal{L}_{\text{sup}}$, whereas the decoder gradient comes from $\mathcal{L_{\text{full}}}$. We verify this hypothesis in section 4.1.

In terms of model architecture, our TM is a standard transformer-based seq2seq model with copy attention \cite{Gu2016IncorporatingCM} (illustrated in Fig.\ \ref{fig:architecture} of \ref{ap_sec:architecture_exp_details}). We fine-tune BERT as the encoder and use a $4$-layer transformer decoder. 
There is little SP-specific inductive bias in the architecture. The only special structure is the copy attention, which is not a strong inductive bias designed for SP as copy attention is widely used in other tasks as well.

% We used a shared WordPiece tokenization for the source natural language and target programming language which makes the use of target auto-encoding simple. Besides, \citet{lample2018phrase} showed that shared byte-pair encoding is helpful if there are shared tokens in different languages.

We refer to the method of using copied monolingual data and freezing the encoder over them as {\em target autoencoding} (TAE). Unless otherwise specified in the ablation studies, the encoder is always frozen.

\vspace{-0.15cm}
\section{Experiments}
\vspace{-0.2cm}
For our primary experiments we considered two python datasets namely Django and CoNaLa. The former is based on Django web framework and the latter is annotated code snippets from stackoverflow answers. Additionally, we experiment on the SQL version of GeoQuery and ATIS from \citet{finegan2018improving} (with query split), WikiSQL \cite{zhong2017seq2sql}, and Magic (Java) \cite{ling2016latent}.

\textbf{Python Monolingual Corpora:} CoNaLa comes with $600K$ mined questions from Stackoverflow. We ignored the noisy source intents/sentences and just use the python snippets. To be comparable with \citet{xu2020incorporating}, we also select a corresponding $100K$ subset version for comparison. See Appendix \ref{sec:appen_data} for details on the SQL and Java monolingual corpora.

\textbf{Experimental Setup:} In all experiments, we use label smoothing with a parameter of $0.1$ and Polyak averaging \cite{polyak1992acceleration} of parameters with a momentum of $0.999$ except for GeoQuery which we use $0.995$. We use Adam \cite{kingma2014adam} and early stopping based on the dataset specific evaluation metric on dev set. The learning rate for the encoder is $1\times10^{-5}$ over all datasets. We used the learning rate of $7.5\times10^{-5}$ on all datasets except GeoQuery and ATIS which we use $1\times10-4$. The architecture overview is shows in Fig.\ \ref{fig:architecture}. 
At the inference time we use beam search with beam size of $10$ and a length normalization based on \cite{Wu2016GooglesNM}. We run each experiment with $5$ different random seeds and report the average and standard deviation. WordPiece tokenization is used for both natural language utterances and programming code.

\vspace{-0.1em}
\subsection{Empirical Analysis}
\vspace{-0.1em}
\label{sec:ablation}

\begin{table*}[t!]
 \begin{minipage}{.5\textwidth}
\centering 
\begin{tabular}{|p{.15\linewidth}|p{.74\linewidth}|}
\hline
{\tiny \textbf{Source:}} & {\tiny \textbf{
\makecell{call the function lazy with 2 arguments : \_string\_concat and \\ six.text\_type , substitute the result for string\_concat .
}}} \\
\hline
{\tiny \textbf{Gold \& TAE:}} &  
\begin{lstlisting}[boxpos=b]
string_concat = lazy(_string_concat, six.text_type)
\end{lstlisting} \\ \hline
{\tiny \textbf{Baseline:}} & 
\begin{lstlisting}[boxpos=b]
string_concat = lazy (<@\textcolor{red}{_concat_}@>concat , six.text_type )
\end{lstlisting}
%% \\ \hline
%% {\tiny TAE:} & 
%% \begin{lstlisting}[boxpos=b]
%% string_concat = lazy ( _string_concat , six.text_type )
%% \end{lstlisting}
 \\\hline
{\tiny \textbf{Note:}} & {\tiny \textbf{copy mistake: wrong variable resulting from failed copy} }\\
 \hline
 %%%%%%%%%%%%%%%%%%%%%%%%%%%%%%%%%%%%%%%%%%%%%%%%%%%%%%%
\end{tabular}
%\captionof*{table}{Pass-through mistake examples}%\label{table:pass_through_examples}
 \end{minipage}\hfill
 \begin{minipage}{.5\textwidth}
%\end{table}
%\begin{table}[t]
\centering 
\begin{tabular}{|p{.15\linewidth}|p{.76\linewidth}|}
\hline
{\tiny \textbf{Source:}} & {\tiny \textbf{
\makecell{
define the function timesince with d , now defaulting to none,\\  reversed defaulting to false as arguments .
}}} \\
\hline
{\tiny \textbf{Gold \& TAE:}} &  
\begin{lstlisting}[boxpos=b]
def timesince(d, now=none, reversed=false):
    pass
\end{lstlisting} \\ \hline
{\tiny \textbf{Baseline:}} & 
\begin{lstlisting}[boxpos=b]
def timesince <@\textcolor{red}{(d = none, reversed ( d = false )}@>:
    pass 
\end{lstlisting}
%% \\ \hline
%% {\tiny TAE:} & 
%% \begin{lstlisting}[boxpos=b]
%% def timesince ( d, now = none, reversed = false ) :
%%     pass 
%% \end{lstlisting}
 \\ \hline
{\tiny \textbf{Note:}} & {\tiny \textbf{unbalanced paranthesis and multiple semantic mistakes.}}\\
\hline
 %%%%%%%%%%%%%%%%%%%%%%%%%%%%%%%%%%%%%%%%%%%%%%%%%%%%%%%
\end{tabular}
%\captionof*{table}{Grammar or semantic mistake examples}%\label{table:grammar_mistake_examples}
%\end{table}
 \end{minipage}
 \caption{\label{table:mistake_examples_short} Example mistakes by the baseline that are fixed by TAE. More examples in Appendix \ref{ap_sec:more_examples_appendix}.}
\end{table*}

\begin{figure}
    \centering
    \includegraphics[width=\linewidth]{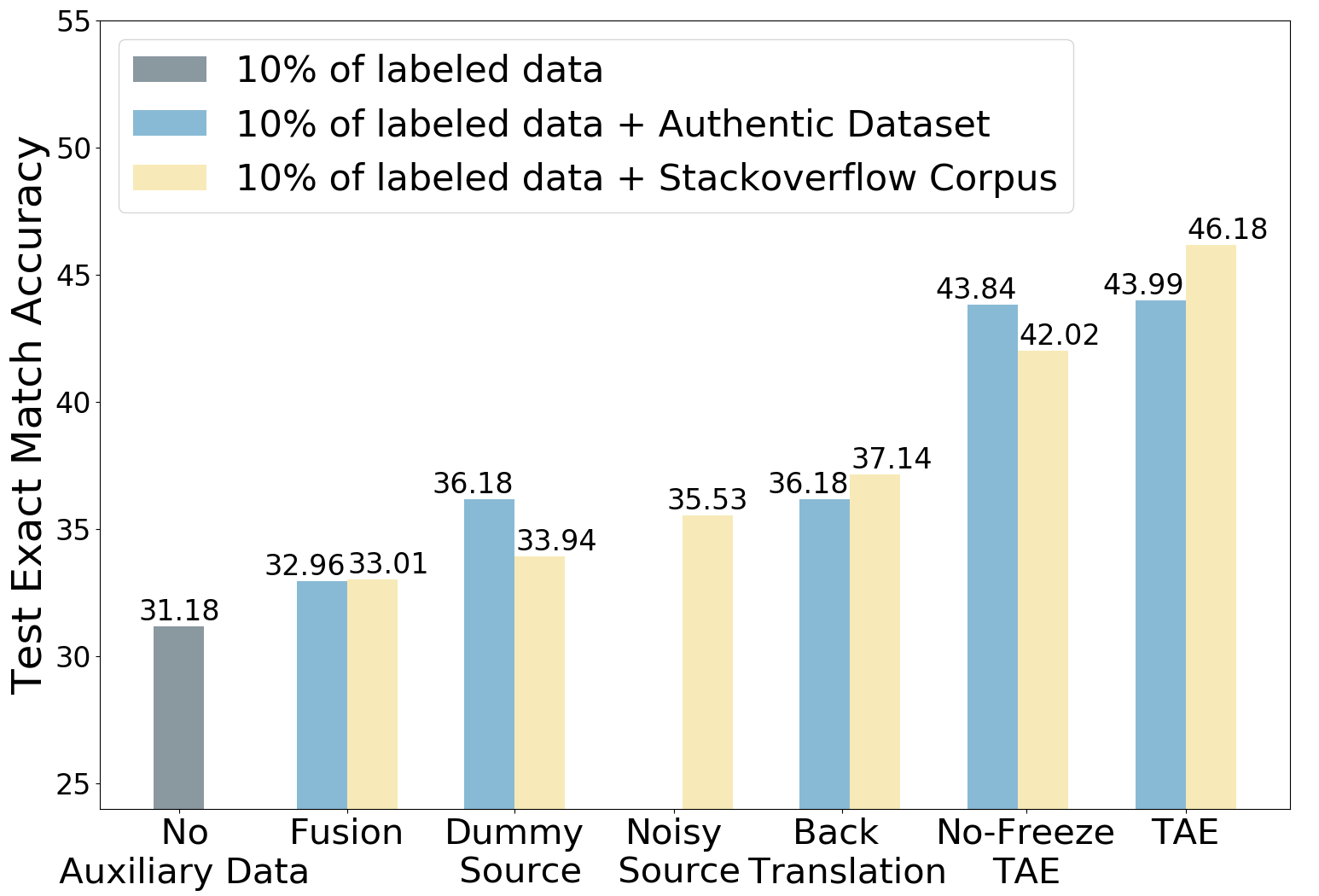}
    \vspace{-1em}
    \caption{Analysis using only 10\% Django train bitext.}
    \label{fig:semi-supervised-django}
\vspace{-.5em}
\end{figure}

First, we considered a scenario where the monolingual corpus comes from the same distribution as the bitext. We simulate this setup by using $10\%$ of Django training data as labeled data while using all the python examples from Django as the monolingual dataset of $10$ times bigger. Results with ``Authentic Dataset'' in Fig.\ \ref{fig:semi-supervised-django} shows the effectiveness of TAE vs other approaches.

Next, we used the monolingual dataset prepared for python (StackOverflow Corpus) which is from a different distribution. Fig.\ \ref{fig:semi-supervised-django}  shows even more considerable improvement, thanks to the larger monolingual set. We considered noisy intents provided in CoNaLa monolingual corpus and dummy source sentences where each monolingual sample is paired along with a random length array containing zeros. We also compared against other well-known approaches like fusion and back-translation, see experiments details in Appendix \ref{ap_sec:fusion}. TAE outperforms all those approaches by a large margin.

Now one important question is, what part of the model benefits from monolingual data most? In Sec.\ \ref{sec:main_method}, we conjectured that auto-encoding of monolingual data should mostly help the decoder, not the encoder. To verify this, we perform an ablation by comparing freezing encoder parameters versus not freezing over the monolingual set. Fig.\ \ref{fig:semi-supervised-django} shows that without freezing the encoder, performance drops slightly for TAE on authentic Django while dropping significantly when copying on Stackoverflow data. This confirms that the performance gain is due to its effect on the decoder, while the copied monolingual data might even hurts the encoder.

%So in later experiments, we always keep the encoder frozen on the monolingual set.

% One might also ask whether the copied target as source is necessary or not i.e.\ what if the auxiliary training is conditioned on some {\em dummy source} instead of the target code. To verify, we mask the encoder inputs with the value of zero for the auxiliary training data. Fig.\ \ref{fig:semi-supervised-django} shows the dummy source is less effective. We also considered using noisy intents as source sentence, and figure \ref{fig:semi-supervised-django} indicates its inferior performance. 
%% \simon{This confused me.  When you say "source sentence"  you actually mean "monolingual target sentence" that will be copied over.  Also, I am assuming that you are replacing some of the tokens with "NULL" at random, but the way you have written is suggests that you are replacing with all NULL tokens.  Finally, tell me what the experiment shows!}
%% \sajad{it's the latter one. replacing all the tokens with NULL}

%% Finally, Fig.\ \ref{fig:semi-supervised-wikisql} shows that similar improvements transfers to SQL domain as well. The only difference is that improvements from the external monolingual dataset are smaller compared to authentic data. This is not surprising given that our SQL monolingual set is five times smaller than WikiSQL's own authentic monolingual set.

\vspace{-0.1em}
\subsection{Main Results on Full Data}
\vspace{-0.1em}

Table \ref{table:django}-\ref{table:conala} showcase our SOTA results on Django and CoNaLa. While our simple base seq2seq model does not outperform previous works, with TAE on the monolingual data, our performance improves and outperforms all the previous works. 
% One important observation here is that the domain of our monolingual dataset is different from the Django dataset, but it still provides helpful signals for the model.

The most direct comparison is with \citet{xu2020incorporating} that also leverage the same extra data mined from StackOverflow ({\em EK} in Table \ref{table:conala}). As mentioned in Sec.\ \ref{sec:related}, they used the noisy parallel corpus for pre-training, whereas we only leverage the monolingual set. However, we obtain both larger relative improvements over our baseline ($32.29$ from $30.98$) compared to \citet{xu2020incorporating} ($28.14$ from $27.20$), as well as better absolute results in the best case. In fact, with only the $100K$ StackOverflow monolingual data, our result is on par with the best one from \citet{xu2020incorporating} that uses the additional python API bitext data. Finally, note that part of our superior performance is due to using BERT as an encoder.

% \subsection{Results on Other Data}
% To examined the effectiveness of TAE for semantic parsing, we considered other three natural language to source code generation datasets: WikiSQL and Geography are SQL queries with corresponding questions; Magic contains Java classes describing card games. The experiment setup details are described in Appendix D. Table.\ref{table:bleu} and Table.\ref{table:exact-match} show the quantitative results. The improvements from TAE on WikiSQL and Magic Dataset are within the model variation. However, when we only utilized $10\%$ labeled training data, TAE achieved significant improvements on both WikiSQL and Magic Dataset. 

\begin{table}[h]
\centering
\small
\begin{tabular}{lll}
\hline \textbf{Model} & \textbf{Django} \\ \hline
\small YN17 \scriptsize \cite{yin2017syntactic} & $71.6$ \\
\small TRANX \scriptsize \cite{yin2018tranx} & $73.7$ \\
\small Coarse2Fine \scriptsize \cite{dong2018coarse} & $74.1$\\
\small TRANX2 \scriptsize \cite{yin2019reranking} & $77.3 \pm 0.4$ \\
\small TRANX2 + BERT & $79.7 \pm 0.42$ \\
\small Reranker \scriptsize \cite{yin2019reranking}$^*$ & $80.2 \pm 0.4$ \\
\hline
\small Our baseline & $77.05 \pm 0.6$ \\
\small Our baseline + TAE &  {\boldmath $ 81.03 \pm 0.14$} \\
\hline
\end{tabular}
\vspace{-0.5em}
\caption{\label{table:django} Exact match accuracy for Django test set. \citet{yin2019reranking}$^*$ trained a separate model on top of SP to rank beam search outputs.}
    \vspace{1em}
\centering
\small
\begin{tabular}{lll}
\hline \textbf{Model} & \textbf{CoNaLa} \\ \hline
\small Reranker \scriptsize \cite{yin2019reranking}$^*$ & $30.11$ \\
\small TRANX {\scriptsize \cite{yin2019reranking}} + BERT & $30.47 \pm 0.7$ \\
\small EK (baseline) \scriptsize \cite{xu2020incorporating} & $27.20$ \\
\small EK + 100k \scriptsize \cite{xu2020incorporating} & $28.14$ \\
\small EK + 100k + API \scriptsize \cite{xu2020incorporating}$^{*\dagger}$ & $32.26$ \\
\hline
\small Our baseline & $30.98 \pm 0.1$ \\
\small Our baseline + TAE on 100k &  $ 32.29 \pm 0.4$ \\
\small Our baseline + TAE on 600k &  {\boldmath $ 32.57  \pm 0.3$} \\
\hline
\end{tabular}
\vspace{-0.5em}
\caption{\label{table:conala} CoNaLa test BLEU. Methods with $^{*}$ trained a separate model on top of SP to rerank beam search outputs. \citet{xu2020incorporating}$^{\dagger}$ used an additional bitext corpus mined from python API documentation.}
    \vspace{1em}
   \centering
       \small
    \begin{tabular}{ccc}
\hline \textbf{Dataset} & \textbf{Baseline (\%)} & \textbf{Baseline + TAE (\%)} \\ \hline
GeoQuery & $ 47.69 \pm 0.05$ & $51.87 \pm 0.02$ \\
ATIS & $38.04 \pm 0.77$ & $ 40.56 \pm 0.57$ \\
Magic  & $41.61 \pm 2.07$ &  $42.34 \pm 0.52$ \\
WikiSQL  & $85.36 \pm 0.06$  &$85.30 \pm 0.07$ \\
\hline
\end{tabular}
\vspace{-0.5em}
\caption{\label{table:extra_dataset_result} Additional dataset results: test set exact match accuracy on all dataset.}
\vspace{-0.5em}
\end{table}

Finally, TAE also yields improvements on other programming languages, as shown for GeoQuery (SQL), ATIS (SQL) and Magic (Java) in Table \ref{table:extra_dataset_result}. We observe no improvement on WikiSQL. But it is not surprising given its large dataset size and the simplicity of its targets. As observed by previous works \cite{finegan2018improving}, more than half of queries follow simple pattern of ``{\fontfamily{qcr}\selectfont SELECT col FROM table WHERE col = value}".

The main results in terms of improvement over previous best methods are statistically significant in Table \ref{table:django}-\ref{table:conala}. On Django, our result is better than Reranker \cite{yin2019reranking} (best previous method in Table \ref{table:django}) with a P-value $< 0.05$, under one-tailed two-sample t-test for mean equality. Since the previous state of the art on CoNaLa (EK + 100k + API in Table \ref{table:conala}) did not provide the standard deviation, we cannot conduct a two-sample t-test against it. Instead, we performed a one-tailed two-sample t-test against the TranX+BERT baseline and observed that our improvement is statistically significant with P-value $< 0.05$. In Table \ref{table:extra_dataset_result}, improvements on GeoQuery and ATIS are statistically significant with P-value $< 0.05$, while it is not the case for Magic and WikiSQL.

\subsection{Discussion}
Thus far, we have verified that the decoder benefits from TAE and the encoder does not. For a better understanding of what TAE improves in the decoder, we propose two metrics namely \textit{copy accuracy} and \textit{generation accuracy}. Copy accuracy only considers tokens appearing in the source sentence. If the model produces all of the tokens that need to be copied from the source sentence, and in the right order, then the score is one otherwise zero for the example. Generation-accuracy ignores tokens appearing in the source intent and computes the exact match accuracy of the prediction. We show how to compute these metrics for the following example:
\\
\textbf{Question:} define the function timesince with d, now defaulting to none, reversed defaulting to false as arguments.
\\
\textbf{Ground Truth:}\\
``{\fontfamily{qcr}\selectfont
def timesince(d, now=none, reversed=false): pass}''

We iterate over the ground truth script tokens one by one and remove those that can be copied from the source, leading to this code:\\
\textbf{Generation Ground Truth:} \\
``{\fontfamily{qcr}\selectfont def (=none=):pass}'',
and the removed tokens will be considered for copy ground truth.\\
\textbf{Copy Ground Truth:} ``{\fontfamily{qcr}\selectfont timesince d , now , reversed false}''.

We would then use the copy and generation ground truth strings to compute each metric. Note that the order of tokens are still important and exact equality is required.

% in \ref{table:examples}. For ignoring the tokens appearing in  . 
As shown in Table \ref{table:accuracies} both metrics are improved. Table \ref{table:mistake_examples_short} illustrates one example from each type and with more samples in the Appendix \ref{ap_sec:more_examples_appendix}. Copy accuracy is important for producing the right variable names mentioned, and it is improved as expected. It is also encouraging to see quantitatively and qualitatively that grammar mistakes are reduced, meaning that the lack of prior knowledge of target language structure is compensated by learning from monolingual data.
%% Up to this point we verified that the decoder benefits from TAE and encoder does not. To have a better understanding of what gets better in the decoder, we propose two metrics namely \textit{copy accuracy} and \textit{generation accuracy}. Copy accuracy only considers tokens that are appearing in the source sentence i.e. if the model reconstruct all of the tokens need to be coppied from the source sentence with the right order then it receives score. Generation-accuracy ignores tokens appearing in the source intent and computes exact match accuracy based on newly generated tokens. As shown in \ref{table:accuracies} both metrics shows improvements. Table \ref{table:mistake_examples_short} depicts one example from each set and more samples provided in the Appendix \ref{ap_sec:more_examples_appendix}. Copy accuracy is especially important in semantic parsing as the model needs to remember the attribute and function names. Besides, generation accuracy shows that model has a better understand of the target language structure.
\begin{table}[h]
    \centering
    \small
    \begin{tabular}{l|c|c}
        \hline
         Model & Copy & Generation \\
         \hline
         10\% basline & 34.18 & 55.73 \\
         10\% baseline + TAE & 58.89 & 66.31 \\
         Full baseline & 80.11 & 81.27 \\
         Full baseline + TAE & 84.59 & 82.65\\
         \hline
    \end{tabular}
\vspace{-0.5em}
    \caption{\label{table:accuracies} {\small Copy and generation accuracies on Django test set} }
\vspace{-1em}
\end{table}

\section{Conclusion}
\vspace{-0.15cm}
% This work has shown a promising alternative direction for making future progress on semantic parsing, and points to the need for better SP evaluation datasets and better SQL monolingual corpus in future research.
This work has shown the possibility to achieve a competitive or even SOTA performance on semantic parsing with little or no inductive bias design. Besides the usual large-scale pre-trained encoders, the key is to exploit relatively large monolingual corpora of the meaning representation. The modified copied monolingual data approach from machine translation literature works well in this extremely low-resource setting. Our results point to a promising alternative direction for future progress.

\section*{Acknowledgements}
We appreciate the ACL anonymous reviewers and area chair for their valuable inputs. We also would like to thank a number of Borealis AI colleagues for helpful discussions, including Wei (Victor) Yang, Peng Xu, Dhruv Kumar, and Simon J.\ D.\ Prince for feedback on the writing.

\bibliography{acl2021}
\bibliographystyle{acl_natbib}

\newpage
\appendix

\onecolumn

\section{Datasets}
\label{sec:appen_data}

We used 6 datasets in total. Django includes programs from Django web framework and CoNaLa contains diverse set of intents annotated on python snippets gathered from Stackoverflow. WikiSQL, GeoQuery, and ATIS include natural language questions and their corresponding SQL queries. WikiSQL includes single table queries while GeogQuery and ATIS requires queries on more than one table. Finally, Magic has Java class implementation of game cards with different methods used during the game. Table \ref{table:datasets_para} summarises all the parallel datasets. For GoeQuery we used query split provided by \cite{finegan2018improving}.
\vspace{0.1cm}

\textbf{Monolingual Corpus}: 
CoNaLa comes with $600K$ mined questions from Stackoverflow. We ignored the noisy source intents/sentences and just use the python snippets. To be comparable with \citet{xu2020incorporating}, we also select a corresponding $100K$ subset version for comparison.
For SQL, \citet{yao2018staqc} automatically parsed StackOverflow questions related to SQL and provided a set containing $120K$ SQL examples. We automatically parsed the SQL codes and removed samples with grammatical mistakes. We also filtered samples not starting with SELECT special token.
\citet{githubCorpus2013} downloaded full repositories of individual projects that were forked at least once; duplicate projects were removed. We randomly sampled 100K Java examples from more than 14K projects and use that as monolingual set. Table \ref{table:datasets_mono} summarises all the monolingual datasets.

\begin{table}[h]
\centering 
% \scriptsize
\begin{tabular}{ccccc}
\hline \textbf{Parallel Corpus} & \textbf{Language} & \textbf{Train} & \textbf{Dev} & \textbf{Test}\\ \hline
Django {\tiny \cite{oda2015ase:pseudogen1}  \href{https://github.com/odashi/ase15-django-dataset/tree/master/django}{(link)}}  & Python & $16000$ & $1000$ & $1805$	 \\
CoNaLa {\tiny \cite{yin2018mining} \href{https://conala-corpus.github.io/}{(link)}} & Python & $2,179$ & $200$ & $500$ \\
WikiSQL {\tiny \cite{zhong2017seq2sql} \href{https://github.com/salesforce/WikiSQL}{(link)}} & SQL & $56,355$  & $8421$ & $15878$\\
ATIS {\tiny \cite{finegan2018improving} \href{https://github.com/jkkummerfeld/text2sql-data/blob/master/data/atis.json}{(link)}}  & SQL & 4812 & 121 & 347\\
GeoQuery {\tiny \cite{finegan2018improving} \href{https://github.com/jkkummerfeld/text2sql-data/blob/master/data/geography.json}{(link)}}  & SQL & 536 & 159 & 182\\
Magic {\tiny \cite{ling-etal-2016-latent} \href{https://github.com/deepmind/card2code/tree/master/third_party/magic}{(link)}} & Java & $8,457$ &  $446$ & $483$\\
\hline
\end{tabular}
\caption{\label{table:datasets_para} Parallel dataset sizes. We filtered out Magic data with java code longer than 350 tokens in order to fit in GPU memory.}
\end{table}

\begin{table}[h]
\centering 
% \scriptsize
\begin{tabular}{ccccc}
\hline
\end{tabular}

\begin{tabular}{ccc}
\hline \textbf{Monolingual Corpus} & Source & \textbf{Size} \\ \hline
Python {\tiny \cite{yin2018mining} \href{https://conala-corpus.github.io/}{(link)}}  & Stackoverflow  & 100K \\
SQL {\tiny \cite{yao2018staqc} \href{https://github.com/LittleYUYU/StackOverflow-Question-Code-Dataset}{(link)}} & Stackoverflow  & 52K \\
Java {\tiny \cite{githubCorpus2013} \href{http://groups.inf.ed.ac.uk/cup/javaGithub/}{(link)}}& Github  & 100k\\
\hline
\end{tabular}
\caption{\label{table:datasets_mono} Monolingual dataset sizes.}
\end{table}

\section{Dev Set Results}

% For each dataset, table \ref{table:exact-match} reports the corresponding evaluation metric on our baseline and TAE augmented model. We conjecture that TAE improves the performance whenever monolingual data provides better coverage over dev or test set. To validate our hypothesis, we use the corpus BLEU score of test queries as a quantitative measurement of how well training data covers test query patterns. 
% We measured the relative corpus BLEU score improvement from adding monolingual data and reported this value in table \ref{table:exact-match}). This metric validates the fact that whenever monolingual data doesn not change the test set coverage we should not expect improvements out of TAE.

\begin{table}[h]
\centering 

\begin{tabular}{cccc}
\hline \textbf{Dataset} & \textbf{Baseline (\%)} & \textbf{Baseline + TAE (\%)} \\ \hline
CoNaLa & $32.43 \pm 0.21$ & $ 34.81 \pm 0.36$ \\
% & $4.23$ &  $24.52$\\
ATIS & $ 5.79 \pm 0.29$ & $7.23 \pm 0.45$ \\
GeoQuery & $ 53.33 \pm 1.47$ & $52.58 \pm 0.70$ \\
% & $8.76$ &  $0.87$\\
Django  & $75.52 \pm 0.21$ & $78.56 \pm 0.39$ \\
% & $5.17 $ &  $29.57$	 \\
Magic  & $42.26 \pm 1.42$ &  $44.17 \pm 0.99$ \\
% & $1.75$ &  $4.80$\\ 
WikiSQL  & $85.92 \pm 0.09$  &$85.83 \pm 0.07 $ \\
% & $-0.70$ &  $0.59$\\

\hline
\end{tabular}

\caption{\label{table:exact-match} Dev set exact match accuracy on all datasets except CoNaLa which uses BLEU. We followed \cite{yin2018tranx} implementation of BLEU score which can be found \href{https://github.com/pcyin/tranX/blob/master/datasets/conala/bleu_score.py}{here}.} 
\end{table}

% \begin{table}[!h]
% \centering 

% \begin{tabular}{ccccc}
% \hline \textbf{Dataset} & \textbf{Data Usage (\%) } & \textbf{Baseline} & \textbf{Baseline + TAE}  & \textbf{${\Delta}${Coverage} (\%)} \\ \hline
% CoNaLa & $100$ & $30.98\pm0.10$ & $32.29\pm0.40$  &  $28.33$ \\
% Django  & $100$ & $59.39 \pm 0.09$ & $61.49 \pm 0.07$ &  $26.84$	 \\
% WikiSQL & $10$  & $75.65 \pm 0.12$  &$
% 81.05 \pm 0.18
% $ &  $11.33$ \\
% Magic & $10$ & $78.33 \pm 0.13$ &  $81.02 \pm 0.45$ & $7.316$\\ 
% Magic & $100$ & $94.88 \pm 0.15$ &  $95.05 \pm 0.04$  & $4.89$ \\ 
% GeoQuery & $100$  & $74.93 \pm 0.28$ & $77.91 \pm 0.53$ &  $1.03$\\
% WikiSQL & $100$  & $92.95 \pm 0.02$  &$93.06 \pm 0.03$  &  $0.54$\\

% \hline
% \end{tabular}

% \caption{\label{table:bleu} BLEU on test set. Experiments are sorted by $\Delta$Coverage. TAE made significant improvements in shaded rows.}
% \end{table}

\section{Architecture and Experiment Details}
\label{ap_sec:architecture_exp_details}
 We selected the decoder learning rate based on linear search over [$1\times10^{-3} -2.5\times10^{-5}$]. Number of decoder layers has been decided based on search over \{2, 3, 4, 5, 6\} layers and 4 layer decoder shows superior performance (we used a single run for hyperparameter selection). Each model has 150M parameters optimized using a single GTX 1080 Ti GPU. With batch size of 16 each step takes ~1.7s on GeoQuery dataset (other datasets have very similar runtime).
 On Django and CoNaLa, we followed \cite{yin2018tranx, xu2020incorporating} on replacing quoted values with a ``str\#'' where \# is a unique id. On Magic dataset, we replaced all newline ``\textbackslash n'' tokens with ``\#''; following \cite{ling-etal-2016-latent}, we splitted Camel-Case words (e.g., class TirionFordring $\rightarrow$ class Tirion Fordring) and all punctuation characters. We filtered out Magic data with java code longer than 350 tokens in order to fit in GPU memory.

% There is little SP-specific inductive bias in the architecture. The only special structure is the copy attention, which is not a strong inductive bias designed for SP as copy attention is widely used in other tasks as well.
% We used a shared WordPiece tokenization for the source natural language and target programming language which makes the use of target auto-encoding simple. Besides, \citet{lample2018phrase} showed that shared byte-pair encoding is helpful if there are shared tokens in different languages.

% As our main purpose is to compare the model performance with and without target auto-encoding, we threw away Java programs longer than 350 tokens in both parallel and monolingual dataset in order to fit in GPU memory.

\section{Back-Translation and Fusion details}
\label{ap_sec:fusion}
\begin{figure}[h]
    \centering
    \includegraphics[width=0.6\linewidth]{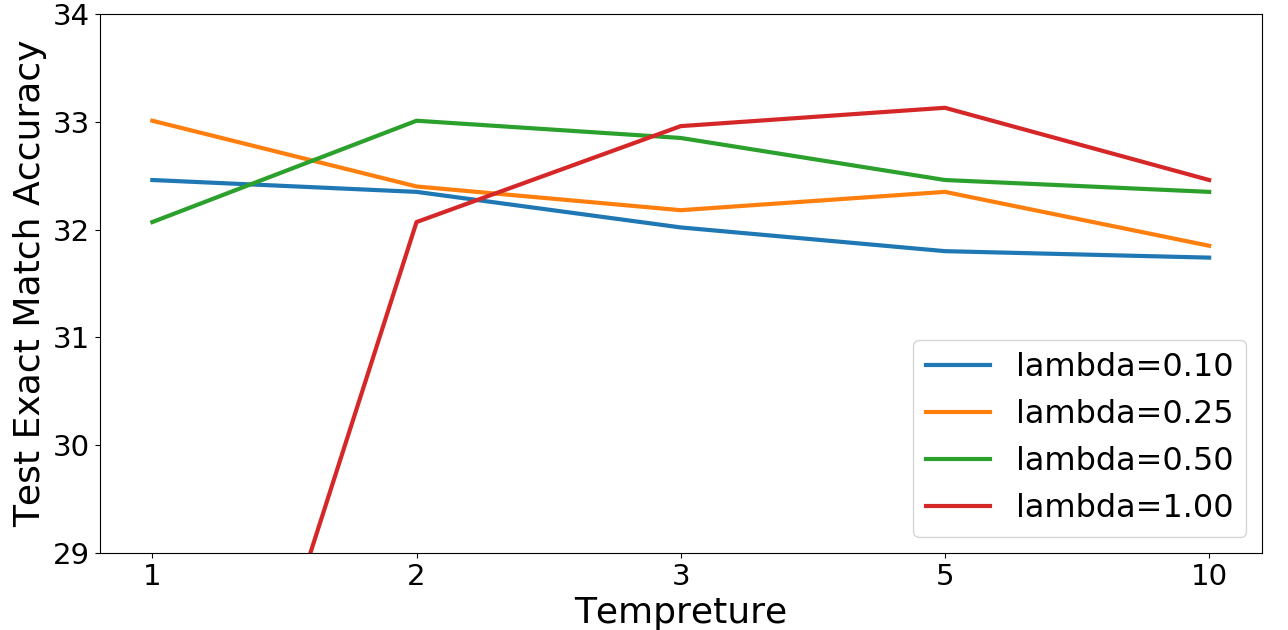}
    \caption{Test exact match accuracy of TM leverage fusion with different parameters}
    \label{fig:fusion}
\end{figure}

For fusion we follow equation \ref{eq:fusion} where TM stands for translation model and LM stands for language model. $\tau$ limits the confidence of the language model and $\lambda$ controls the balance between TM and LM. figure \ref{fig:fusion} shows the performance of a base TM trained on 10\% of Django training data with test exact match accuracy of $31.80$ over different values of $\lambda$ and $\tau$. The LM is trained over full Django training set.
\begin{equation}
\log p(y^t_i) = \log p_{TM}(y^t_i) + \lambda \log p_{LM}(y^t_i) = \log p_{TM}(y^t_i) + \lambda \log \frac{e^{l_i^t}/\tau}{\sum_i e^{l_i^t}/\tau} 
\label{eq:fusion}
\end{equation}

For back-translation we first trained the model using the same architecture explained above in the backward direction. We used BLEU score as a evaluation metric and use early stopping based on that. Using greedy search we generate the corresponding source intent for each code snippet. In the end, the synthetic data is merged with the bitext and trained a forward model.

\newpage
\section{Additional Qualitative Examples}
\label{ap_sec:more_examples_appendix}

\begin{table*}[!h]
 \begin{minipage}{.5\textwidth}
\centering 
\begin{tabular}{|p{.08\linewidth}|p{.8\linewidth}|}
\hline
{\tiny Source:} & {\tiny \textbf{\texttt{
call the function lazy with 2 arguments : _string_concat and six.text_type [ six . text_type ] , substitute the result for string_concat .
}}} \\
\hline
{\tiny Gold:} &  
\begin{lstlisting}[boxpos=b]
string_concat = lazy(_string_concat, six.text_type)
\end{lstlisting} \\ \hline
{\tiny Baseline:} & 
\begin{lstlisting}[boxpos=b]
string_concat = lazy (<@\textcolor{red}{_concat_}@>concat , six.text_type )
\end{lstlisting}
\\ \hline
{\tiny TAE:} & 
\begin{lstlisting}[boxpos=b]
string_concat = lazy ( _string_concat , six.text_type )
\end{lstlisting}
 \\ \hline
{\tiny Note:} & {\tiny wrong var}\\
 \hline
 %%%%%%%%%%%%%%%%%%%%%%%%%%%%%%%%%%%%%%%%%%%%%%%%%%%%%%%
  \hline
{\tiny Source:} & {\tiny \textbf{\texttt{
get translation_function attribute of the object t , call the result with an argument eol_message , substitute the result for result .
}}} \\
\hline
{\tiny Gold:} &  
\begin{lstlisting}[boxpos=b]
result = getattr(t, translation_function)(eol_message)
\end{lstlisting} \\ \hline
{\tiny Baseline:} & 
\begin{lstlisting}[boxpos=b]
result = getattr ( t , translation_<@\textcolor{red}{message}@> ) ( eol_message ) 
\end{lstlisting}
\\ \hline
{\tiny TAE:} & 
\begin{lstlisting}[boxpos=b]
result = getattr ( t , translation_function ) ( eol_message ) 
\end{lstlisting}
 \\ \hline
{\tiny Note:} & {\tiny wrong var}\\
\hline
 %%%%%%%%%%%%%%%%%%%%%%%%%%%%%%%%%%%%%%%%%%%%%%%%%%%%%%%
\hline
{\tiny Source:} & {\tiny \textbf{\texttt{
convert whitespace character to unicode and substitute the result for space .
}}} \\
\hline
{\tiny Gold:} &  
\begin{lstlisting}[boxpos=b]
space = unicode(' ')
\end{lstlisting} \\ \hline
{\tiny Baseline:} & 
\begin{lstlisting}[boxpos=b]
space = unicode ( <@\textcolor{red}{character}@> ) 
\end{lstlisting}
\\ \hline
{\tiny TAE:} & 
\begin{lstlisting}[boxpos=b]
space = unicode ( ' ' )
\end{lstlisting}
 \\ \hline
{\tiny Note:} & {\tiny wrongly copied variable name}\\
\hline
 %%%%%%%%%%%%%%%%%%%%%%%%%%%%%%%%%%%%%%%%%%%%%%%%%%%%%%%
 \hline
{\tiny Source:} & {\tiny \textbf{\texttt{
assign integer 2 to parts if third element of version equals to zero , otherwise assign it integer 3 .
}}} \\
\hline
{\tiny Gold:} &  
\begin{lstlisting}[boxpos=b]
parts = 2 if version[2] == 0 else 3
\end{lstlisting} \\ \hline
{\tiny Baseline:} & 
\begin{lstlisting}[boxpos=b]
parts <@\textcolor{red}{[ 2 ] = 2}@> 
\end{lstlisting}
\\ \hline
{\tiny TAE:} & 
\begin{lstlisting}[boxpos=b]
parts = 2 if version [ 2 ] == 0 else 3 
\end{lstlisting}
 \\ \hline
{\tiny Note:} & {\tiny baseline failed to copy a few source tokens, and instead formed a grammatically correct but semantically incorrect output}\\
\hline
 %%%%%%%%%%%%%%%%%%%%%%%%%%%%%%%%%%%%%%%%%%%%%%%%%%%%%%%
\end{tabular}
\captionof*{table}{Copy mistake examples}%\label{table:pass_through_examples}
 \end{minipage}
 \begin{minipage}{.5\textwidth}
%\end{table}
%\begin{table}[t]
\centering 
\begin{tabular}{|p{.08\linewidth}|p{.8\linewidth}|}
\hline
{\tiny Source:} & {\tiny \textbf{\texttt{
define the function timesince with d , now defaulting to none , reversed defaulting to false as arguments .
}}} \\
\hline
{\tiny Gold:} &  
\begin{lstlisting}[boxpos=b]
def timesince(d, now=none, reversed=false):
    pass
\end{lstlisting} \\ \hline
{\tiny Baseline:} & 
\begin{lstlisting}[boxpos=b]
def timesince <@\textcolor{red}{( d = none, reversed ( d = false )}@> :
    pass 
\end{lstlisting}
\\ \hline
{\tiny TAE:} & 
\begin{lstlisting}[boxpos=b]
def timesince ( d, now = none, reversed = false ) :
    pass 
\end{lstlisting}
 \\ \hline
{\tiny Note:} & {\tiny unbalanced paranthesis and multiple semantic mistakes.}\\
\hline
 %%%%%%%%%%%%%%%%%%%%%%%%%%%%%%%%%%%%%%%%%%%%%%%%%%%%%%%
\hline
{\tiny Source:} & {\tiny \textbf{\texttt{
define the function exec with 3 arguments : _code_ , _globs_ set to none and _locs_ set to none .
}}} \\
\hline
{\tiny Gold:} &  
\begin{lstlisting}[boxpos=b]
def exec_(_code_, _globs_=none, _locs_=none):
    pass
\end{lstlisting} \\ \hline
{\tiny Baseline:} & 
\begin{lstlisting}[boxpos=b]
def exec ( _code_ , <@\textcolor{red}{_globs}@>= none , <@\textcolor{red}{_locs_ set ( )}@> ) :
    pass 
\end{lstlisting}
\\ \hline
{\tiny TAE:} & 
\begin{lstlisting}[boxpos=b]
def exec ( _code_ , _globs_ = none , _locs_ = none ) :
    pass 
\end{lstlisting}
 \\ \hline
{\tiny Note:} & {\tiny wrong variable name and grammar mistake}\\
\hline
 %%%%%%%%%%%%%%%%%%%%%%%%%%%%%%%%%%%%%%%%%%%%%%%%%%%%%%%
 \hline
{\tiny Source:} & {\tiny \textbf{\texttt{
return an instance of escapebytes , created with an argument , reuslt of the call to the function bytes with an argument s .
}}} \\
\hline
{\tiny Gold:} &  
\begin{lstlisting}[boxpos=b]
return escapebytes(bytes(s))
\end{lstlisting} \\ \hline
{\tiny Baseline:} & 
\begin{lstlisting}[boxpos=b]
return escapebytes ( bytes ( s ) <@\textcolor{red}{. re ( s ) }@>
\end{lstlisting}
\\ \hline
{\tiny TAE:} & 
\begin{lstlisting}[boxpos=b]
return escapebytes ( bytes ( s ) ) 
\end{lstlisting}
 \\ \hline
{\tiny Note:} & {\tiny extra semantically incorrect predictions and unbalanced paratheses}\\
\hline
  %%%%%%%%%%%%%%%%%%%%%%%%%%%%%%%%%%%%%%%%%%%%%%%%%%%%%%%
%% \hline
%% {\tiny Source:} & {\tiny \textbf{\texttt{
%% read data from buf , yield the result .
%% }}} \\
%% \hline
%% {\tiny Gold:} &  
%% \begin{lstlisting}[boxpos=b]
%% yield buf.read()
%% \end{lstlisting} \\ \hline
%% {\tiny Baseline:} & 
%% \begin{lstlisting}[boxpos=b]
%% yield <@\textcolor{red}{buf}@> 
%% \end{lstlisting}
%% \\ \hline
%% {\tiny TAE:} & 
%% \begin{lstlisting}[boxpos=b]
%% yield buf . read ( ) 
%% \end{lstlisting}
%%  \\ \hline
%% {\tiny Note:} & {\tiny semantic mistake}\\
%% \hline
 %%%%%%%%%%%%%%%%%%%%%%%%%%%%%%%%%%%%%%%%%%%%%%%%%%%%%%%
\hline
{\tiny Source:} & {\tiny \textbf{\texttt{
call the function blankout with 2 arguments : p and str0 , write the result to out .
}}} \\
\hline
{\tiny Gold:} &  
\begin{lstlisting}[boxpos=b]
out.write(blankout(p, 'str0'))
\end{lstlisting} \\ \hline
{\tiny Baseline:} & 
\begin{lstlisting}[boxpos=b]
out .write ( blankout ( p , 'str0' <@\textcolor{red}{)}@> 
\end{lstlisting}
\\ \hline
{\tiny TAE:} & 
\begin{lstlisting}[boxpos=b]
out .write ( blankout ( p , 'str0' ) ) 
\end{lstlisting}
 \\ \hline
{\tiny Note:} & {\tiny unbalanced paratheses}\\
\hline 
\end{tabular}
\captionof*{table}{Grammar or semantic mistake examples}%\label{table:grammar_mistake_examples}
%\end{table}
 \end{minipage}
 \caption{\label{table:mistake_examples}Mistake examples}
\end{table*}

% \begin{figure}[h]
%     \centering
%     \includegraphics[width=0.9\linewidth]{figures/semi-supervised-wikisql.png}
%     \caption{Analysis using $10\%$ of WikiSQL training bitext: execution accuracy on WikiSQL dev set}
%     \label{fig:semi-supervised-wikisql}
% \end{figure}

% \section{Experiment Details}

\end{document}